\documentclass[journal]{IEEEtran}
\usepackage{times}%

\usepackage{amssymb}
\usepackage{amsmath}
\usepackage{dcolumn}
\usepackage{graphics}
\usepackage{hyperref}
\usepackage{multirow}
\usepackage{booktabs}
\usepackage{graphicx}
\usepackage{algorithm}
\usepackage{algorithmicx}
\usepackage{algpseudocode}

\begin{document}
\title{ACP: Automatic Channel Pruning via Clustering and Swarm Intelligence Optimization for CNN}

\author{Jingfei~Chang,
        Yang~Lu,
        Ping~Xue,
        Yiqun~Xu,
        and~Zhen~Wei% <-this % stops a space
\thanks{Corresponding author: Yang Lu.}% <-this % stops a space
\thanks{J. Chang, Y. Lu, P. Xue, Y. Xu and Z. Wei are with the School of Computer Science and Information Engineering, Hefei University of Technology, Hefei 230009, China (e-mail: cjfhfut@mail.hfut.edu.cn; luyang@hfut.edu.cn; xueping@mail.hfut.edu.cn; yiqunxu@mail.hfut.edu.cn; weizhen@gocom.cn).}% <-this % stops a space
\thanks{Y. Lu and Z. Wei are also with the Engineering Research Center of Safety Critical Industrial Measurement and Control Technology, Ministry of Education, Hefei University of Technology.}}

\maketitle

\begin{abstract}
As the convolutional neural network (CNN) gets deeper and wider in recent years, the requirements for the amount of data and hardware resources have gradually increased. Meanwhile, CNN also reveals salient redundancy in several tasks. The existing magnitude-based pruning methods are efficient, but the performance of the compressed network is unpredictable. While the accuracy loss after pruning based on the structure sensitivity is relatively slight, the process is time-consuming and the algorithm complexity is notable. In this article, we propose a novel automatic channel pruning method (ACP). Specifically, we firstly perform layer-wise channel clustering via the similarity of the feature maps to perform preliminary pruning on the network. Then a population initialization method is introduced to transform the pruned structure into a candidate population. Finally, we conduct searching and optimizing iteratively based on the particle swarm optimization (PSO) to find the optimal compressed structure. The compact network is then retrained to mitigate the accuracy loss from pruning. Our method is evaluated against several state-of-the-art CNNs on three different classification datasets CIFAR-10/100 and ILSVRC-2012. On the ILSVRC-2012, when removing 64.36\% parameters and 63.34\% floating-point operations (FLOPs) of ResNet-50, the Top-1 and Top-5 accuracy drop are less than 0.9\%. Moreover, we demonstrate that without harming overall performance it is possible to compress SSD by more than 50\% on the target detection dataset PASCAL VOC. It further verifies that the proposed method can also be applied to other CNNs and application scenarios.
\end{abstract}

\begin{IEEEkeywords}
Deep learning, convolutional neural network, channel pruning, model compressing, clustering, swarm intelligence optimization.
\end{IEEEkeywords}

% \fi
\IEEEpeerreviewmaketitle

\section{Introduction}
\IEEEPARstart{W}{ith} the continuous expansion of datasets and the substantial increase in hardware computing power, deep neural network (DNN) models \cite{ISI:000237698100002A01},\cite{ISI:000239308600057A02},\cite{ISI:000355286600030A03} have achieved great success in scientific research and engineering. As one of the noteworthy members, CNN has a better performance in several fields such as image classification \cite{INSPEC:17133663A04}, object detection \cite{ISI:000450913100006A05}, style transfer \cite{ISI:000425498401060A06}, and semantic segmentation \cite{ISI:000380414100170A07} due to the parameter sharing and local connectivity schemes. However, as computer vision tasks become more complex, the depth and width of the network are gradually expanding. While improving performance, the scale of CNN has also become extremely large. As a result, some existing deep CNNs can only be applied in GPU, TPU, or cloud, which considerably limits the development and application of CNNs. Moreover, most researches have revealed that deep CNNs have obvious parameter redundancy in many tasks.

To disentangle the dilemma, many researchers devote themselves to compressing and accelerating the CNNs. The current mainstream methods mainly include designing dedicated hardware \cite{ISI:000360535000019A08}, optimizing convolutional calculation \cite{INSPEC:16989755A09}, and designing network compression schemes. Among them, network compressing is mainly divided into four types: low-rank decomposition \cite{10.1145/3152127A10}, quantification \cite{ISI:000389385100032A11}, knowledge distillation \cite{INSPEC:16231013A12}, and network pruning. The low-rank decomposition can achieve sparseness and directly compress the network. Nevertheless, additional calculations are introduced in the implementation, which is not conducive to the decrease of FLOPs. Quantification can degrade storage requirements and speed up the inference, but it will cause accuracy loss. Knowledge distillation can improve the performance of small networks by learning from the pre-trained large teacher network, however, the performance depends on the similarity of the two tasks. The network pruning is based on the redundancy of CNN and evaluates the importance of the parameters to delete the insignificant filters or channels. Since it is easy to implement and can effectively compress and accelerate the network while maintaining the original accuracy, network pruning has received widespread attention. The existing pruning methods based on the magnitude of filters and channels \cite{ISI:000450913101044A13},\cite{ISI:000425498401048A14},\cite{DBLP:conf/iclr/0022KDSG17A15} or their similarity \cite{DBLP:conf/ijcai/LiMXL20A16}, etc. are relatively efficient, but the accuracy drop of the compact network is unstable. While the pruning schemes via the structure sensitivity can obtain a compressed network with less precision loss, but the pruning phase is time-cost. For example, \cite{DBLP:conf/cvpr/Yu00LMHGLD18A17} considers the influence of error propagation on pruning and discards neurons layer by layer via minimizing the final response layer. \cite{DBLP:conf/eccv/HeLLWLH18A18} implements network pruning through reinforcement learning. In response to the above obstacles, we deliberate to prune the channels for preliminarily compressing and then design an algorithm to further optimize the pruned model. In this way, the performance of the compact network can be improved as much as possible which is even better than the vanilla network. The motivation of our method is two-fold. First, \cite{DBLP:conf/ijcai/LiMXL20A16} shows the effectiveness of leveraging the similarity between feature maps to measure the parameter redundancy. Second, \cite{DBLP:conf/iclr/ZophL17A19} reveals that the neural architecture search can guide to optimize the network pruning.

In this paper, we propose an automatic channel pruning method based on clustering and swarm intelligence optimization. First, we randomly sample and test several images from the dataset using the pre-trained baseline CNN. The feature maps at the same place are integrated to calculate the cosine similarity of channels in each layer. According to the similarity, the feature maps are analyzed and clustered, and the result of the clustering is regarded as the number of remaining channels to produce the preliminary compact network. Then we propose an approach to initialize the network structure population based on the compressed substructure. Finally, the particle swarm optimization is adopted to automatically search for the optimal compact model. The two benefits of performing preliminary clustering pruning and then optimizing the substructure through swarm intelligence searching are as follows. First, the clustering pruning itself is efficient, and it can shrink the searching space for the subsequent iterative optimization, which makes the entire network compressing time-saving. Second, clustering pruning can provide a more reliable initialization for optimization, thereby mitigating the accuracy drop. In this paper, we conduct compressing for VGGNet \cite{DBLP:journals/corr/SimonyanZ14aA20}, ResNet \cite{DBLP:conf/cvpr/HeZRS16A21}, and GoogLeNet \cite{DBLP:conf/cvpr/SzegedyLJSRAEVR15A22} on CIFAR-10, CIFAR-100, and ILSVRC-2012 \cite{DBLP:journals/ijcv/RussakovskyDSKS15A23}, and the SSD \cite{ISI:000389382700002A24} is compressed on PASCAL VOC \cite{ISI:000348345400006A25}.

Extensive experiments indicate that ACP can effectively and accurately compress CNNs. On CIFAR-10, we remove 80.92\% parameters and 78.32\% FLOPs of ResNet-110, and surprisingly gain 0.33\% improvement of performance. While the parameters and FLOPs of GoogLeNet are discarded by 66.29\% and 70.44\% respectively, the accuracy of the compact network is even increased by 0.37\%. For CIFAR-100 with less training images, ACP discards 47.67\% parameters and 52.21\% FLOPs of ResNet-56 with only 0.21\% accuracy drop. On the large-scale ILSVRC-2012, when pruning 54.69\% parameters and 46.82\% FLOPs of ResNet-50, its Top-1 and Top-5 accuracy only reduce 0.41\% and 0.21\%.

The proposed ACP is adapt to all existing CNNs and different tasks. Apart from that, the iterative searching based on PSO is capable of optimizing all channel pruning methods. Moreover, because of no sparseness introducing, ACP does not require the assistance of additional sparse matrix operations and acceleration libraries. And the entire pruning phase can be achieved by controlling only one hyper-parameter, which notably reduces labor intervention and can perform automatic compression.

The main contributions of this paper can be summarized as follows:
\begin{itemize}
	\item We propose a layer-wise clustering pruning method for CNN, which regards the sum of the number of clusters obtained by clustering as the number of channels in each layer.
	\item We introduce a CNN structure optimizing method based on PSO. We propose an initializing scheme to extend the preliminarily pruned network as a structure population, and then apply the PSO to obtain the optimal compact network.
	\item An automatic channel pruning method combined with the clustering and PSO is presented. It can achieve flexible compression for CNN by only adjusting one hyper-parameter under the condition that almost retains or even exceeds the original accuracy.
	\item This paper conducts comprehensive experiments on the image classification datasets CIFAR-10/100, ILSVRC-2012, and further verification on the target detection dataset PASCAL VOC. The results sufficiently confirm the effectiveness of our ACP.
\end{itemize}

\begin{figure*}[htbp]
	
	\centering
	
	\includegraphics[width=17cm]{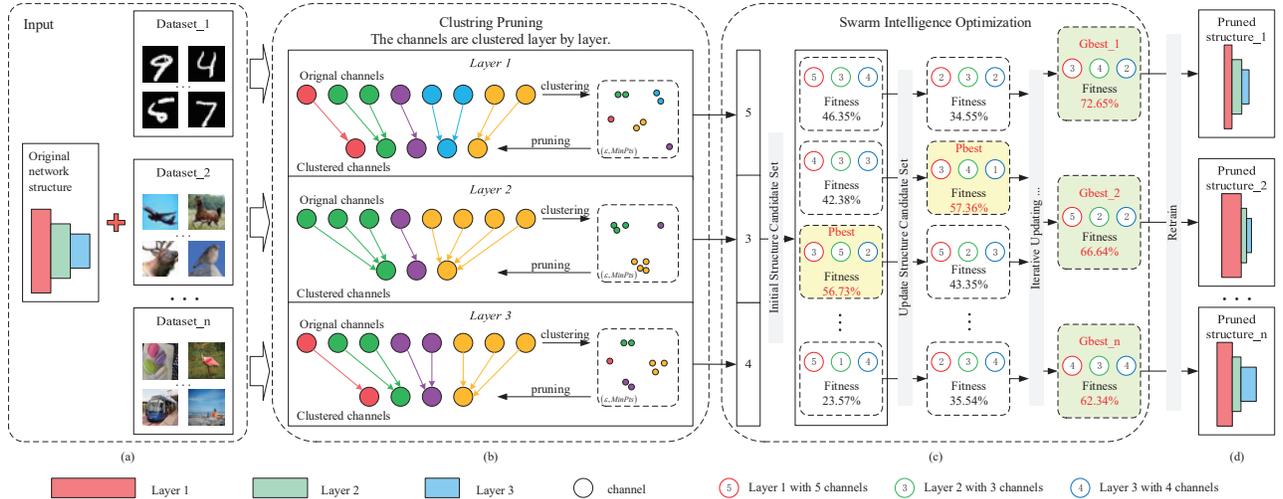} 
	
	\caption{The ACP framework. For simplicity, we take a three-layer CNN as an example. (a) Inputs: original network and different datasets. (b) Clustering Pruning: The redundant channels are clustered and pruned based on the similarity of feature maps. (c) Swarm Intelligence Structure Optimization: The pruned network structure obtained by clustering is initialized as the candidate population. We train a few epochs for each candidate to calculate its fitness and then update all substructures. After some cycles, the substructure with the highest fitness is regarded as the global best compact network. (d) Different datasets end up with various pruned network structures. We finally retrain the pruned model to recover its predictive accuracy. (This figure is best viewed in color and zoomed in.)}
	\label{img1}
	
\end{figure*}

\section{Related Works}

Deep CNNs have made breakthroughs in various fields of computer vision, but their enormous parameters and FLOPs exceedingly limit the application of CNNs. And for some simple tasks, too many parameters will cause over-fitting, which may cause performance deterioration. At present, there have been several methods devoted to network compression and acceleration. Among them, the network pruning algorithm has gained extensive attention due to its simple implementation and apparent effects, and many excellent algorithms have emerged. The existing network pruning algorithms can be roughly divided into two categories, magnitude-based pruning and sensitivity-based pruning.

\subsection{Magnitude-based pruning}
The magnitude-based pruning focuses on the importance of parameters and connections. According to the proposed importance evaluation index, the unimportant parameters and connections in the network are directly deleted, and then the obtained compact network is fine-tuned or retrained to restore the experimental accuracy. The core of this type of algorithms is to design a reasonable and effective parameter importance evaluation index. \cite{ISI:000425498402086A26} uses the scaling factor of the batch normalization layers to evaluate the importance of the channels and deletes the unimportant channels to realize the channel-level network pruning. \cite{ISI:000450913101044A13} uses the magnitude of the weights to measure and deletes all insignificant connections in the pre-trained network whose weight is lower than the threshold. \cite{DBLP:conf/iclr/0022KDSG17A15} calculates and sorts the L1 norm of the filters and prunes them with smaller values and their corresponding feature maps according to the preset pruning rate. \cite{DBLP:conf/ijcai/HeKDFY18A27} utilizes the L2 norm to measure the importance of the filters, and will continuously update the filters that were pruned last time during the training phase. \cite{DBLP:journals/pami/ChenZ19A28} adopts the fully connected layer as a linear classifier to extract the feature representation of the middle layers and prune them that with less improvement according to a predefined threshold. \cite{DBLP:conf/cvpr/HeLWHY19A29} judges the redundancy of one filter based on the Euclidean Distance of it and the geometric mean of all filters in the layer and deletes the redundant filters in the network according to the preset pruning rate. \cite{DBLP:conf/cvpr/LinJWZZ0020A30} applies the rank of the feature map to determine how much information it contains. The low-rank feature maps contain less information and can be deleted with confidence. The high-rank feature maps hold much more message, even if fixing some parameters will not damage the final performance. \cite{DBLP:conf/ijcai/LiMXL20A16} proposes a new filter pruning method, which analyses the diversity and similarity of feature maps to prune redundant filters in the network. \cite{ISI:000553424500017A31} prunes the 3D ConvNets according to the magnitude and information score of the filters.

Clustering is the task of dividing the data points into several groups such that data points in the same groups are more similar to each other than those in other groups. In simple words, the aim is to segregate groups with similar traits and assign them into clusters. Broadly speaking, clustering can be divided into three subgroups: prototype-based clustering, such as k-means algorithm \cite{macqueen1967someA32}, etc.; hierarchical clustering, such as BIRCH algorithm \cite{DBLP:conf/sigmod/ZhangRL96A33}, etc. and density-based clustering, such as DBSCAN algorithm \cite{DBLP:conf/kdd/EsterKSX96A34},\cite{10.1145/3068335A35} and OPTICS Algorithm \cite{DBLP:conf/sigmod/AnkerstBKS99A36}, etc. According to specific magnitude characteristics such as similarity of the parameters, the original network can be clustered and pruned.

%\IEEEpubidadjcol

\subsection{Sensitivity-based pruning}
Sensitivity-based methods require analyzing the influence of pruning on the original network. Concretely, by considering the impact of pruning strategies on network performance in the compressing process, the CNN is pruned layer by layer, and the scheme is adjusted continually during the accelerating phase. The core of this type of algorithm is to design a rational and effective network pruning adjustment strategy. \cite{DBLP:conf/cvpr/Yu00LMHGLD18A17} considers the importance score propagation of the neurons and uses feature ranking techniques to measure the importance of each neuron in the "final response layer" to prune the filters in earlier layers. \cite{DBLP:conf/iclr/LeeAT19A37} introduces connection sensitivity to evaluate the importance of structural connections and prunes the network during the parameter initialization stage before training. \cite{DBLP:conf/iclr/ParkLMS20A38} comprehensively considers the neuron connections before and after the pruned layer and compresses the network by minimizing the formulated Frobenius distortion. \cite{DBLP:conf/cvpr/ZhaoNZZZT19A39} introduces a variational technique to estimate channel saliency in the training process and looks for a suitable probability distribution to further deletes redundant channels. \cite{DBLP:conf/iclr/GaoZDMX19A40} proposes a feature boosting and suppression method to predictively amplify salient convolutional channels and skip unimportant ones at run-time. This method retains the complete network structure. \cite{DBLP:conf/cvpr/HeDLZZ020A41} develops a differentiable pruning criteria sampler that is learnable and optimized by the validation loss of the pruned network obtained from the sampled criteria. So they can adaptively select the appropriate pruning criteria for different functional layers. \cite{DBLP:conf/cvpr/GaoHPH20A42} directly optimizes channel-wise differentiable discrete gate under resource constraint to controls the deletion of channels in the convolutional layers. \cite{DBLP:conf/cvpr/GuoWLY20A43} proposes a differentiable Markov channel pruning method to efficiently compress the network that is differentiable and can be directly optimized by gradient descent with respect to standard task loss and budget regularization. \cite{DBLP:journals/pr/LuoW20A44} proposes an effective channel selection layer, which automatically finds less important filters in a joint training manner. The method takes previous activation responses as an input and generates a binary index code for pruning. \cite{DBLP:journals/ijon/WangLSXW20A45} proposes a network channel pruning scheme based on sparse learning and genetic algorithm.

Swarm intelligence optimization algorithms are a form of nature-based optimization algorithms that mainly simulates the cooperative behavior of insects, birds, and animals, etc. The individuals of the community continuously change the direction to search by sharing knowledge between them until they achieve their goals. Famous swarm intelligence optimization algorithms include ant colony algorithm \cite{INSPEC:4304141A46}, particle swarm optimization algorithm \cite{INSPEC:5297172A47},\cite{699146A48}, and artificial bee colony algorithm \cite{karaboga2005ideaA49}, etc.

\section{Proposed Method}

We propose the ACP to decrease the redundancy of CNN for different tasks and achieve effective and accurate channel-level compression and acceleration by reducing the number of parameters and FLOPs. In this section, we present the overall ACP framework first, and then specifically introduce the two major components: clustering pruning and swarm intelligent structure optimization. Finally, we describe the pruning strategy for different CNNs.

\begin{figure}[t]
	
	\centering
	
	\includegraphics[width=8.5cm]{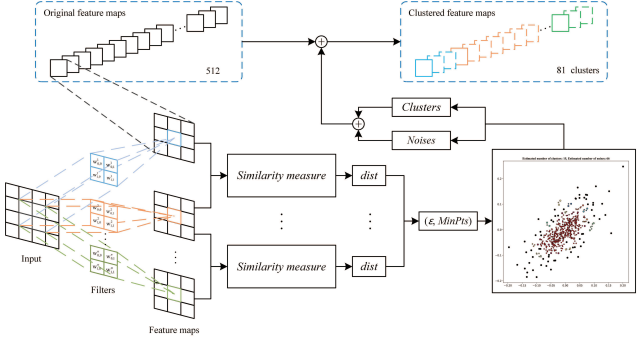} 
	
	\caption{Illustration of clustering pruning.}
	\label{img2}
	
\end{figure}

\subsection{The Automatic Channel Pruning Framework}

Fig.\ref{img1} shows the overall framework of ACP. Given an image classification dataset, we randomly sample numerous images and infer them via the pre-trained CNN. The feature maps are clustered based on the density and the number of clusters in each layer is regarded as the number of channels after pruning. Then the obtained network substructure is initialized as a structure candidate population. According to the accuracy loss of the subnetwork, we iteratively search and optimize each candidate using PSO. Finally, after a few cycles, we can obtain the optimal compact structure via automatic channel pruning. For different tasks and datasets, the compressed network structure is generally different.

\subsection{Clustering Pruning}

Given a CNN with \textit{L} convolutional layers and the dataset $D$, the convolutional operation is formulated as $Conv = {{\cal F}}\left( {D,W;C} \right)$. $C = \left( {{c_1},{c_2}, \ldots ,{c_L}} \right)$ is the original network structure, where ${c_l}$ is the number of channels in the $l$-th layer. The size of the filter $W\in {{\mathbb{R}}^{{{c}_{out}}\times {{c}_{in}}\times k\times k}}$ is $k \times k$, where ${c_{out}}$ and ${c_{in}}$ denote the number of output and input channels, respectively. The feature map is a 4D tensor $M\in {{\mathbb{R}}^{s\times c\times W\times H}}$ with the size of $W \times H$, where $s$ is the number of samples from the dataset.

Here, we perform layer-wise clustering pruning based on the similarity of feature maps to achieve the preliminary network compressing. The number of groups formed by clustering is regarded as the number of channels after pruning. Given an input image, the feature map generated in the $l$-th layer is ${{M}_{l}}\in {{\mathbb{R}}^{{{c}_{l}}\times {{W}_{l}}\times {{H}_{l}}}}$, and then the ${{c}_{l}}$ channels of which the size is ${{W}_{l}}\times {{H}_{l}}$ are clustered. We draw on the idea of the Density-Based Spatial Clustering of Applications with Noise (DBSCAN) algorithm. The density-based clustering can separate regions with adequate density into clusters with arbitrary shapes in the dataset containing noise. Moreover, there is no need to specifically assign the number of classes to be formed before clustering.

Different filters perform convolutional operations on the input to generate various feature maps, and some of them in the same layer may have similarities. For a given dataset containing $m$ training images, we randomly pick $s$ $\left( s\le m \right)$ samples and process them through the CNN. To fully consider the impact of different types of figures on the output feature maps and obtain more accurate similarity, we superimpose and average the feature maps at the same position produced by $s$ samples, and then calculate the cosine distance to estimate the similarity between any two feature maps ${{M}_{li}}$ and ${{M}_{lj}}$ in the same layer. The specific definition is as follows:

\begin{equation}
{dist({M_{lij}}) = \left| {\cos \left[ {\left( {\frac{1}{s}\sum\limits_{r = 1}^s {M_{li}^r} } \right),\left( {\frac{1}{s}\sum\limits_{r = 1}^s {M_{lj}^r} } \right)} \right]} \right|},
\label{eq1}
\end{equation}

where $M_{li}^{r}$ is the $i$-th feature map produced by the $r$-th image in the $l$-th layer. The neighborhood parameter of the cluster is ($\varepsilon$,$MinPts$), where $\varepsilon$ is the neighborhood radius, and $MinPts$ refers to the minimum number of feature maps in the $\varepsilon$-neighborhood of the core feature map. If the $\varepsilon$-neighborhood of the feature map ${{M}_{li}}$ contains at least $MinPts$ feature maps, that is $\left| {{\mathbb{N}}_{\varepsilon }}\left( {{M}_{li}} \right) \right|\ge MinPts$, then ${{M}_{li}}$ is assigned as a core feature map, and all of the core feature maps form a set. We randomly select a core feature map and cluster all feature maps including non-core ones in its $\varepsilon$-neighborhood into a group. If there is a core feature map, then all the feature maps in its $\varepsilon$-neighborhood also belong to this group. Feature maps that are neither core feature maps nor belong to $\varepsilon$-neighborhood of any core feature maps are regarded as noises. According to this scheme, we cluster the channels in each layer and consider the sum of the number of groups and noises obtained by clustering as the number of channels after clustering pruning. In this way, the primary compressed substructure ${C}'=\left( {{c}_{1}^{\prime}},{{c}_{2}^{\prime}},\ldots ,{{c}_{L}^{\prime}} \right)$ is gained. Fig.\ref{img2} depicts the process of clustering and pruning in one layer. The upper right corner shows the substructure after pruning, and the feature maps in the same color belong to the same cluster. The number of feature maps in the solid line indicates the number of remaining channels.

\subsection{Swarm Intelligence Structure Optimization}

In the previous section, we have conducted primary clustering pruning for the original network structure $C$. In this section, the compressed substructure ${C}'$ is further optimized through structure searching. We use the PSO algorithm to optimize ${C}'$ and find the optimal compact network ${{C}^{*}}$ which satisfies:

\begin{equation}
{{{C}^{*}}=\underset{{{C}'}}{\mathop{\arg \max }}\,Acc\left( {{D}_{Test}}\left( {{D}_{Train}},{W}';{C}' \right) \right)}
\end{equation}

where ${{D}_{Train}}$ denotes training set and ${{D}_{Test}}$ is test set. $Acc\left(\centerdot\right)$ is the test accuracy of the structure ${C}'$. The structure searching optimization can be divided into the following two steps.

\subsubsection{Network Structure Initialization}

First of all, we initialize the particle swarm according to the compressed structure ${C}'$ and obtain a structure candidate population set $\left\{ C_{n}^{0} \right\}$ with $N$ substructure candidates, where $n\in \left[ 1,N \right]$. In order to achieve locally random initialization that makes the initialized network structure similar to ${C}'$ but different in several layers, we propose a structure population initialization scheme. The specific implementation is as follows:

\begin{equation}
{c_{nl}^{0}={{c}_{l}}^{\prime }+n\times rand\left\{ -1,0,1 \right\}}
\end{equation}

where the $n$-th structure is $C_{n}^{0}=\left( c_{n1}^{0},c_{n2}^{0},\ldots ,c_{nL}^{0} \right)$, and $rand\left\{ -1,0,1 \right\}$ denotes sampling a number from -1, 0 and 1 randomly. Then, we initialize the searching velocity set $\left\{ V_{n}^{0} \right\}$ of channels, where $V_{n}^{0}=\left( v_{n1}^{0},v_{n2}^{0},\ldots ,v_{nL}^{0} \right)$, and the max velocity is set to ${{v}_{max}}$ so that $v_{n}\in \left[ -{{v}_{max}},{{v}_{max}} \right]$. The fitness of $C_{n}^{0}$ can be calculated via Eq.\ref{eq4}.

\begin{equation}
{fitness\left( {{C}_{n}} \right)=Acc\left( {{D}_{Test}}\left( {{D}_{Train}},{{W}_{n}};{{C}_{n}} \right) \right)}
\label{eq4}
\end{equation}

Since the network training is time-consuming, it is cumbersome to calculate fitness many times during the swarm intelligence structure searching and optimization. To tackle this problem, we only train a few epochs for each structure to obtain fitness. To do so, it can not only reveals the performance of the structure candidate but also saves the searching time. The $N$ network structures are initialized as the local best structure $pbes{{t}_{n}}$ respectively and the structure with the greatest fitness is regarded as the global best network $gbest$.

\subsubsection{Iterative Searching}

After the initialization, iterative searching is performed to find the optimal CNN structure. First, we update the searching velocity as follows:

\begin{equation}
\begin{split}
v_{n}^{t}=&w\times v_{n}^{t-1}+{{\alpha }_{1}}\times rand\times \left( pbes{{t}_{n}}-c_{n}^{t-1} \right)\\
&+{{\alpha }_{2}}\times rand\times \left( gbest-c_{n}^{t-1} \right)
\end{split}
\label{eq5}
\end{equation}

where $v_{n}^{t}$ denotes the channel searching velocity of the $n$-th candidate in the $t$-th iteration. ${{\alpha }_{1}}$ and ${{\alpha }_{2}}$ are two learning weight that are two positive constants. In this paper, we adopt the recommend value ${{\alpha }_{1}}={{\alpha }_{2}}=2$ in \cite{699146A48}, since it on average makes the weights for “social” and “cognition’’ parts to be 1. $rand$ is a random number between 0 and 1, while $w$ is inertia weight. The dynamic $w$ behave better in the searching optimization, therefore, we utilize the linear decreasing weight scheme for $w$ as follow:

\begin{equation}
{{{w}^{t}}={\left( {{w}_{max}}-{{w}_{min}} \right)\left( T-t \right)}/{T}\;+{{w}_{min}}}
\end{equation}

where ${{w}_{max}}$ denotes the initial inertia weight, while ${{w}_{min}}$ is the inertia weight at the final iteration. Here, ${{w}_{max}}={0.9}$ and ${{w}_{min}}={0.4}$. $T$ is the maximum iteration cycle and $t$ denotes the current cycle. Then we update the candidate $C_{n}^{t}$ as follows:

\begin{equation}
{c_{n}^{t}=c_{n}^{t-1}+r\times v_{n}^{t}}
\label{eq7}
\end{equation}

where $r$ is the learning rate of channels and we set $r=2$. After that, we calculate the fitness of each candidate. If it is greater than that of the initial local best structure, we assign the candidate as the new $pbes{{t}_{n}}$. Furthermore, if it is larger than that of the global best network, $gbest$ is replaced by the candidate. In the end, after iteratively searching for $T$ cycles, the $gbest$ is the optimal pruned structure ${{C}^{*}}$. Algorithm \ref{alg1} manifests the pseudocode of our ACP. Given a pre-trained network, we retrain the obtained compact network from pruning and optimizing to mitigate the accuracy loss from compressing.

\begin{algorithm}[htbp]
	\caption{Automatic Channel pruning (ACP)}
	\label{alg:algorithm}
	\textbf{Input}: Original structure: $C$, Cycles: $T$, Number of initialization structures: $N$. \\
	\textbf{Output}: Optimal pruned structure: ${C}^{*}$. 
	\begin{algorithmic}[1] %[1] enables line numbers
		\State Calculate the cosine similarity between feature maps in each layer via Eq.\ref{eq1};
		\State Cluster the feature maps in each layer to obtain the pruned structure ${C}'$;
		\For{$n=1$ to $N$}  
		\State Initialize the pruned structure set $\left\{ C_{n}^{0} \right\}$;
		\State Initialize the searching velocity set $\left\{ V_{n}^{0} \right\}$;
		\State Calculate $fitness\left( C_{n}^{0} \right)$ of ${{C}_{n}^{0}}$ via Eq.\ref{eq4};
		\State $pbes{{t}_{n}}=C_{n}^{0}$;
		\EndFor 
		\State $gbest=\max \{pbes{{t}_{n}}\}$;
		\For{$t=1$ to $T$} 
		\For{$n=1$ to $N$}  
		\State Update the searching velocity $V_{n}^{t}$ via Eq.\ref{eq5};
		\State Update the candidate $C_{n}^{t}$ via Eq.\ref{eq7};
		\State Calculate $fitness\left( C_{n}^{t} \right)$ of candidate ${C_{n}^{t}}$ via Eq.\ref{eq4};     
		\If {$fitness\left( C_{n}^{t} \right)>fitness\left( pbes{{t}_{n}} \right)$}  
		\State $pbes{{t}_{n}}=C_{n}^{t}$;
		\State $fitness\left( pbes{{t}_{n}} \right)=fitness\left( C_{n}^{t} \right)$;  
		\EndIf
		\If {$fitness\left( pbes{{t}_{n}} \right)>fitness\left( gbest \right)$}  
		\State $gbest=pbes{{t}_{n}}$;
		\State $fitness\left( gbest \right)=fitness\left( pbes{{t}_{n}} \right)$;  
		\EndIf
		\EndFor
		\EndFor
	\end{algorithmic}
	${C}^{*}=gbest$.
	\label{alg1}
\end{algorithm}

For most of the existing network pruning methods, the compressed network inherits the weights and bias from the original network to restore the performance as much as possible through fine-tuning. However, when the network pruning rate is remarkable, the accuracy recovery after fine-tuning is not obvious, and the actual performance of the compact network cannot be greatly manifested. \cite{DBLP:conf/iclr/LiuSZHD19A50} makes a surprising observation in structured network pruning that fine-tuning a pruned model only gives comparable or worse performance than training that model with randomly initialized weights. And the experiment results reveal that the pruned architecture itself, rather than a set of inherited “important” weights, is more crucial to the efficiency in the final model. The results of pruning VGG-16 on CIFAR-10 further verify the observation in \cite{DBLP:conf/iclr/LiuSZHD19A50}. In order to fully demonstrate the performance of the compact network, we retrain them from scratch in our experiments. Specifically, we keep the number of FLOPs consistent before and after pruning. The number of training epochs of the original network multiplied by the accelerating rate of FLOPs is regarded as the retraining epochs. Finally, we compare the accuracy of the pruned network with the original network to draw a conclusion.

\subsection{Pruning Strategy for Different CNNs}

\begin{figure}[t]
	
	\centering
	
	\includegraphics[width=8.5cm]{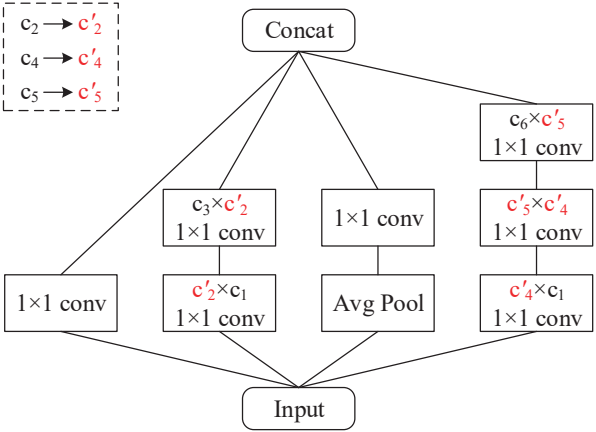} 
	
	\caption{Illustration of pruning Inception V3 module. The black font indicates the number of original channels, and the red font indicates that after pruning.}
	\label{img3}
	
\end{figure}

Since different CNNs have different network structures, different pruning schemes need to be adopted for different networks to maintain structural stability. We perform experiments on VGGNet, ResNet, and GoogLeNet respectively. Among these CNNs, VGGNet is a common layer-by-layer CNN. During the compressing process, the output channels of each layer can be deleted independently. The number of input channels of each layer remains the same as the output of the previous layer. ResNet and GoogLeNet leverage the Residual and Inception module to make the network deeper and wider respectively. If the channels are discarded arbitrarily, the dimensional matching of the network will be destroyed. For ResNet with the basic block, we only prune the output of the first layer, and the input of the second layer is changed accordingly. The bottleneck consists of three layers, among which the size of the filters in the first and third layers is 1$\times$1, while that in the middle layer is 3$\times$3. We prune the channels in the middle layer of the bottleneck. And the output channels in the first layer and the input channels of the third layer will be modified accordingly. GoogLeNet is a more complex network, composed of multiple Inception V3 modules that contain four branches. To prune the GoogLeNet, we compress the branches containing two and three convolutional layers. The specific structure and pruning scheme of the Inception V3 module is plotted in Fig.\ref{img3}.

\section{Experiments}

\subsection{Implementation Details}

In order to demonstrate the effectiveness of our proposed method, we prune VGGNet, ResNet, and GoogLeNet on CIFAR-10, CIFAR-100, and ILSVRC-2012 for image classification. Moreover, we prune SSD on PASCAL VOC for object detection. We implement all experiments using PyTorch. The network pre-training and hyper-parameter settings use the method presented in \cite{DBLP:conf/cvpr/HeZRS16A21} for sake of obtaining better benchmark accuracy and be able to compare fairly with some existing network pruning methods. The learning rate is initially set to 0.1 and then decreased by a factor of 10 on half and three-quarter epochs. We train 160 epochs on CIFAR-10 and CIFAR-100 and 100 epochs on ILSVRC-2012. During training, all the networks are trained using SGD, and batch size is set to be 64 on CIFAR-10/100 and 16 on ILSVRC-2012. We use a weight decay of $1e$-$4$ and a momentum of 0.9. In the retraining stage, we adjust the learning rate as described above on the CIFAR while adopting the cosine annealing adjustment strategy on the ILSVRC-2012. We only change the neighborhood radius   in the clustering pruning to control the compression ratio of the networks. Other hyper-parameter settings are as follows: $MinPts$=5, $T$=5, $N$=6. We perform specific ablation analysis on $\varepsilon$ and MinPts later. We train each compact network for three epochs before calculating its fitness.

\begin{table}[t]	
	\centering
	\caption{Accuracy and pruning ratio of VGG-16, ResNet-56/110 and GoogLeNet on CIFAR-10. $``$Acc.drop$"$ is the accuracy drop of the pruned network, so a negative number means the compressed model has better performance than the baseline. A smaller number of $``$Acc.drop$"$ is better. $``$Parameters.drop$"$ and $``$FLOPs.drop$"$ are the pruned percentage of the parameters and FLOPs, respectively.}
	\renewcommand{\arraystretch}{1.3}
	\resizebox{85mm}{!}{
		\begin{tabular}{c c c c c c c}			
			\hline 
			Model            & Acc/\% & Acc.drop/\%  & Parameters/M & Parameters.drop/\% & FLOPs/M & FLOPs.drop/\% \\ 
			\hline 
			\textbf{VGG-16 Baseline}           & 93.60  & $-$          & 14.73        & $-$                & 314.59  & $-$           \\ 	
			ACP($\varepsilon$=0.010)  & 94.03  & -0.43        & 5.02         & 65.72              & 152.64  & 51.48           \\ 
			ACP($\varepsilon$=0.020)  & 93.66  & -0.06        & 2.76         & 81.28              & 93.52   & 70.25           \\
			ACP($\varepsilon$=0.035)  & 93.45  & 0.15         & 1.23         & 91.66              & 83.54   & 73.44           \\
			\hline
			\textbf{Resnet-56 Baseline}        & 93.18  & $-$          & 0.85         & $-$                & 127.62  & $-$           \\ 	
			ACP($\varepsilon$=0.035)  & 93.78  & -0.60        & 0.59         & 30.59              & 78.31   & 38.64           \\ 
			ACP($\varepsilon$=0.075)  & 93.39  & -0.21        & 0.35         & 58.82              & 58.17   & 54.42           \\
			ACP($\varepsilon$=0.080)  & 92.91  & 0.27         & 0.30         & 64.71              & 53.80   & 57.84           \\
			\hline
			\textbf{Resnet-110 Baseline}       & 93.32  & $-$          & 1.73         & $-$                & 256.04  & $-$           \\ 	
			ACP($\varepsilon$=0.085)  & 94.33  & -1.01        & 0.65         & 62.43              & 92.29   & 63.95           \\ 
			ACP($\varepsilon$=0.130)  & 94.07  & -0.75        & 0.49         & 71.68              & 78.11   & 69.49           \\
			ACP($\varepsilon$=0.300)  & 93.65  & -0.33        & 0.33         & 80.92              & 55.52   & 78.32           \\
			\hline
			\textbf{GoogLeNet Baseline}        & 94.72  & $-$          & 6.17         & $-$                & 1533.96 & $-$          \\ 	
			ACP($\varepsilon$=0.030)  & 95.16  & -0.44        & 3.95         & 35.98              & 901.60   & 41.22        \\ 
			ACP($\varepsilon$=0.070)  & 95.12  & -0.40        & 2.90         & 53.00              & 620.36   & 59.56        \\
			ACP($\varepsilon$=0.200)  & 95.09  & -0.37        & 2.08         & 66.29              & 453.38   & 70.44        \\		
			\hline 			
	\end{tabular}}	
	\label{tab1}	
\end{table}

\begin{table}[t]	
	\centering
	\caption{Accuracy and pruning ratio of VGG-19 and ResNet-56/110 on CIFAR-100.}
	\renewcommand{\arraystretch}{1.3}
	\resizebox{85mm}{!}{
		\begin{tabular}{c c c c c c c}			
			\hline 
			Model            & Acc/\% & Acc.drop/\%  & Parameters/M & Parameters.drop/\% & FLOPs/M & FLOPs.drop/\% \\ 
			\hline 
			\textbf{VGG-19 Baseline}  & 72.01  & $-$          & 20.09        & $-$                & 399.52  & $-$           \\ 	
			ACP($\varepsilon$=0.005)  & 73.12  & -1.10        & 10.03        & 50.07              & 239.77  & 39.99           \\ 
			ACP($\varepsilon$=0.010)  & 71.95  & 0.06         & 5.22         & 74.02              & 146.87  & 63.24           \\
			\hline
			\textbf{Resnet-56 Baseline} & 71.36  & $-$        & 0.86         & $-$                & 127.09  & $-$           \\ 	
			ACP($\varepsilon$=0.030)  & 71.77  & -0.41        & 0.58         & 32.56              & 82.21   & 35.31           \\ 
			ACP($\varepsilon$=0.060)  & 71.15  & 0.21         & 0.45         & 47.67              & 60.73   & 52.21           \\
			\hline
			\textbf{Resnet-110 Baseline} & 72.15  & $-$       & 1.73         & $-$                & 256.04  & $-$           \\ 	
			ACP($\varepsilon$=0.040)  & 73.00  & -0.85        & 0.96         & 44.51              & 135.71   & 47.00           \\ 
			ACP($\varepsilon$=0.070)  & 72.49  & -0.34        & 0.74         & 57.23              & 119.33   & 53.39           \\			
			\hline 			
	\end{tabular}}	
	\label{tab2}	
\end{table}

\begin{table*}[t]	
	\centering
	\caption{Accuracy and pruning ratio of ResNet-18/34/50 on ILSVRC-2012.}
	\renewcommand{\arraystretch}{1.3}
	\resizebox{\textwidth}{!}{
		\begin{tabular}{c c c c c c c c c}			
			\hline 
			Model  &Top-1 Acc/\% & Top-1 Acc.drop/\% & Top-5 Acc/\% & Top-5 Acc.drop/\% & Parameters/M & Parameters.drop/\% & FLOPs/M & FLOPs.drop/\% \\ 
			\hline 
			\textbf{Resnet-18 Baseline}  & 70.02  & $-$       & 89.23        & $-$      &   11.69    &  $-$     & 1819.87  & $-$     \\ 	
			ACP($\varepsilon$=0.005)  & 67.82  & 2.20         & 88.41        & 0.82     &   5.06     &  56.72   & 1197.99  & 34.17   \\ 
			ACP($\varepsilon$=0.010)  & 66.24  & 3.78         & 87.12        & 2.11     &   4.51     &  61.42   & 959.66   & 47.27   \\
			\hline
			\textbf{Resnet-34 Baseline} & 73.40  & $-$        & 91.44        & $-$      &   21.80    &  $-$     & 3672.07  & $-$     \\ 	
			ACP($\varepsilon$=0.010)  & 73.03  & 0.37         & 91.32        & 0.12     &   15.38    &  29.45   & 2525.73  & 31.22   \\ 
			ACP($\varepsilon$=0.030)  & 72.08  & 1.32         & 90.57        & 0.87     &   9.14     &  58.07   & 1625.64  & 55.73   \\
			\hline
			\textbf{Resnet-50 Baseline} & 75.94  & $-$       & 92.93         & $-$      &   25.56    &  $-$     & 4112.32  & $-$   \\ 	
			ACP($\varepsilon$=0.020)  & 75.53  & 0.41        & 92.72         & 0.21     &   11.58    &  54.69   & 2186.87  & 46.82  \\ 
			ACP($\varepsilon$=0.040)  & 74.64  & 0.89        & 92.19         & 0.53     &   9.11     &  64.36   & 1507.46  & 63.34  \\			
			\hline 			
	\end{tabular}}	
	\label{tab3}	
\end{table*}

\subsection{Results on CIFAR-10}

We first prune VGG-16 and ResNet-56/110 on CIFAR-10. The results are shown in TABLE \ref{tab1}. It can be seen that for the three networks when pruning more than half of the parameters and FLOPs, the performance of all the compact networks even has a slight improvement compared with the vanilla network. For VGG-16, we remove 65.72\% parameters and 51.48\% FLOPs while still keeping the accuracy at 94.03\%, which is even 0.43\% higher than the baseline. When the parameters and FLOPs of ResNet-56 are discarded by 58.82\% and 54.42\%, respectively, the precision is improved by 0.21\%. While the pruning rates of parameters and FLOPs for ResNet-110 are 62.43\% and 63.95\%, the performance of the compact network surprisingly increased by 1.01\%. Generally speaking, with the increase of the compression ratio, the accuracy of the pruned model drops gradually. Our method reduces up to 91.66\% parameters and 73.44\% FLOPs for VGG-16 on CIFAR 10 while maintaining the accuracy as high as 93.45\%, which is only declined by 0.15\% compared with the original network. We remove 64.71\% parameters and 57.84\% FLOPs for ResNet-56 with only 0.27\% performance drop of accuracy. For ResNet-110, although compressing 80.92\% parameters and 78.32\% FLOPs, the performance of the network is still improved by 0.33\% compared to the original network.

The experimental results show that the deep CNN does have certain redundancy, and the moderate compression will not affect its accuracy. When the pruning ratio is small, removing some redundant parameters in the network can improve its generalization, which leads to an increase in performance. As ResNet-56 is compressed to 0.59M, the classification performance on CIFAR-10 even exceeds the original VGG-16 of 14.73M. It also supports the excellent performance of the residual structure in image feature extraction and classification tasks from the side. When ResNet-110 is pruned to 0.33M, the parameters and FLOPs are both less than the pruned ResNet-56 of 0.35M, but the accuracy is 0.26\% ahead. The result sheds light on that increasing the depth of the ResNet can improve its performance when the number of parameters of networks are similar.

To further validate the effectiveness of the proposed ACP, we also perform experiments on GoogLeNet. Due to the integration of the Inception module, GoogLeNet increases the width of the network, and finally, the baseline accuracy reach 94.72\%, which is ahead of VGGNet and ResNet. We remove 66.29\% parameters and 70.44\% FLOPs while still holding the accuracy at 95.09\%, which is even 0.37\% higher than the baseline model. It further manifests that the proposed method can achieve effective compression and acceleration for CNN. And the ACP also plays the role of regularization while reducing the redundancy of the network.

\subsection{Results on CIFAR-100}

We continue to experiment with the CNN on CIFAR-100 and the results are reported in TABLE \ref{tab2}. CIFAR-100 and CIFAR-10 have the same number of images, but the category has increased from 10 to 100. As the training images of each class decrease, the performance of the network also drops significantly. The initial classification accuracy of VGG-19 on CIFAR-100 is only 72.01\%. When pruning 74.02\% parameter and 63.24\% FLOPs, the accuracy only drops by 0.06\%. We remove 57.23\% parameters and 53.39\% FLOPs of ResNet-110 while even 0.34\% higher than the baseline model. Under the circumstance, the parameter and FLOPs of 0.74M ResNet-110 are both smaller than the 0.86M original ResNet-56, but the performance is 1.13\% higher than the ResNet-56. The result further verifies the conclusion in the previous section.

\subsection{Results on ILSVRC-2012}

To sufficiently demonstrate the effectiveness of our ACP, we prune ResNet-18/34/50 on the large-scale dataset ILSVRC-2012 that contains 1000 categories. Among them, ResNet-18/34 use basic block, while ResNet-50 uses bottleneck. We arrange two pruning ratios for each network, and the results are presented in TABLE \ref{tab3}. We can see that the accuracy of all the three networks after compressing dose have a loss compared with the original network, moreover as the pruning ratio increases the accuracy loss also raises. It reveals that the three ResNets have less parameter redundancy on ILSVRC-2012. In spite of this, our ACP can still obtain flexible compression and acceleration with slight accuracy drop. When delete 56.72\% parameters and 34.17\% FLOPs of ResNet-18, the Top-1 and Top-5 accuracy drop by 2.20\% and 0.82\%, respectively. For ResNet-34, the initial Top-1 and Top-5 accuracy are 73.40\% and 91.44\%. When the pruning rate of the parameters and FLOPs exceeds 50\%, the Top-1 and Top-5 accuracy only drop 1.32\% and 0.87\%. Because of the bottleneck, the Top-1 and Top-5 accuracy of ResNet-50 increase by 2.54\% and 1.49\% respectively, while the parameters only increase by 3.76M compared to ResNet-34. We remove 54.69\% parameters and 46.82\% FLOPs of ResNet-50, and the performance of the compact network is hardly reduced (0.41\% Top-1 loss and 0.21\% Top-5 loss). When we discard 64.36\% parameters and 63.34\% FLOPs, the compact ResNet-50 are smaller than the baseline ResNet-18, but the accuracy is significantly improved (Top-1: 74.64\% versus 70.02\%, Top-5: 92.19\% versus 89.23\%).

\subsection{Comparison with Other Methods}

In this subsection, we compare the method on three datasets including CIFAR-10, CIFAR-100, and ILSVRC-2012. Among the compared methods, Li et al. \cite{DBLP:conf/iclr/0022KDSG17A15}, SFP \cite{DBLP:conf/ijcai/HeKDFY18A51}, DCP \cite{DBLP:conf/nips/ZhuangTZLGWHZ18A52}, FPGM \cite{DBLP:conf/cvpr/HeLWHY19A29}, EDP \cite{9246734A53}, CNN-FCF \cite{DBLP:conf/cvpr/LiWYFZL19A54}, CCP \cite{DBLP:conf/icml/PengWCH19A55}, Taylor-FO-BN \cite{DBLP:conf/cvpr/MolchanovMTFK19A56} and HRank \cite{DBLP:conf/cvpr/LinJWZZ0020A30} are the state-of-the-art channel pruning methods. The results of these competing methods are reported according to the original article.

\begin{table}[t]
	\centering
	\caption{Performance comparison of ResNet-56 on CIFAR-10. The $``$-$"$ indicates that the results are not listed in the original article.}
	\renewcommand{\arraystretch}{1.3}
	\resizebox{85mm}{!}{
		\begin{tabular}{c c c c c c}			
			\hline 
			Method  &Baseline Acc/\% & Pruned Acc/\% & Acc.drop/\% & Parameters.drop/\% & FLOPs.drop/\% \\ 
			\hline
			SFP \cite{DBLP:conf/ijcai/HeKDFY18A51}         & 93.59  & 93.89  & -0.30    & $-$    & 14.70   \\
			Li et al. \cite{DBLP:conf/iclr/0022KDSG17A15}   & 93.04  & 93.06  & -0.02    & 13.70    & 27.60   \\
			Liu et al. \cite{DBLP:conf/iclr/LiuSZHD19A50}   & 93.14  & 93.05  &  0.09    & 13.70   & 27.60   \\
			HRank \cite{DBLP:conf/cvpr/LinJWZZ0020A30}       & 93.26  & 93.52  & -0.26  & 29.30   & 16.80   \\
			\textbf{ACP($\varepsilon$=0.035)}  & \textbf{93.18}  & \textbf{93.78}  & \textbf{-0.60}  & \textbf{30.59} & \textbf{38.64}   \\
			\hline
			\hline
			SFP \cite{DBLP:conf/ijcai/HeKDFY18A51}         & 93.59  & 93.78  & -0.19   & $-$   & 41.10   \\
			HRank \cite{DBLP:conf/cvpr/LinJWZZ0020A30}       & 93.26  & 93.17  &  0.09  & 50.00  & 42.40   \\
			CNN-FCF \cite{DBLP:conf/cvpr/LiWYFZL19A54}     & 93.14  & 93.38  & -0.24  & 43.09  & 42.78   \\
			NISP \cite{DBLP:conf/cvpr/Yu00LMHGLD18A17}        & $-$    & $-$    &  0.03   & 42.60  & 43.61   \\			
			Y.He et al. \cite{DBLP:conf/cvpr/HeDLZZ020A41}  & 93.59  & 93.72  & -0.13    & $-$   & 47.10   \\
			He et al. \cite{ISI:000425498401048A14}   & 92.80  & 91.80  &  1.00  & $-$   & 50.00   \\			
			AMC \cite{DBLP:conf/eccv/HeLLWLH18A18}         & 92.80  & 91.90  &  0.90  & $-$ & 50.00   \\
			DCP \cite{DBLP:conf/nips/ZhuangTZLGWHZ18A52}         & 93.80  & 93.59  &  0.31  & $-$   & 50.00   \\			
			DMC \cite{DBLP:conf/cvpr/GaoHPH20A42}         & 93.62  & 93.69  & -0.07  & $-$   & 50.00   \\			
			SFP \cite{DBLP:conf/ijcai/HeKDFY18A51}         & 93.59  & 93.35  &  0.24   & $-$    & 52.60   \\		
			FPGM \cite{DBLP:conf/cvpr/HeLWHY19A29}        & 93.59  & 92.89  &  0.70   & $-$ & 52.60   \\			
			CCP \cite{DBLP:conf/icml/PengWCH19A55}         & 93.50  & 93.42  &  0.08  & $-$  & 52.60   \\			
			Y.He et al. \cite{DBLP:conf/cvpr/HeDLZZ020A41}  & 93.59  & 93.34  &  0.25  & $-$  & 52.90   \\			
			ABCPruner \cite{DBLP:conf/ijcai/LinJZZW020A57}   & 93.26  & 93.23  &  0.03   & 54.20  & 54.13   \\			
			EDP \cite{9246734A53}         & 93.61  & 93.61  &  0      & 54.18     & 57.71   \\
			\textbf{ACP($\varepsilon$=0.075)}  & \textbf{93.18}  & \textbf{93.39}  & \textbf{-0.21} & \textbf{58.82} & \textbf{54.42}   \\			
			\hline 			
	\end{tabular}}	
	\label{tab4}	
\end{table}

\begin{table}[t]	
	\centering
	\caption{Performance comparison of ResNet-110 on CIFAR-10.}
	\renewcommand{\arraystretch}{1.3}
	\resizebox{85mm}{!}{
		\begin{tabular}{c c c c c c}			
			\hline 
			Method  &Baseline Acc/\% & Pruned Acc/\% & Acc.drop/\% &  Parameters.drop/\%  & FLOPs.drop/\% \\ 
			\hline
			SFP \cite{DBLP:conf/ijcai/HeKDFY18A51}          & 93.68  & 93.83  & -0.15  & $-$    & 14.60   \\
			Li et al. \cite{DBLP:conf/iclr/0022KDSG17A15}   & 93.53  & 93.55  & -0.02  & 2.30   & 15.90   \\
			Liu et al. \cite{DBLP:conf/iclr/LiuSZHD19A50}  & 93.14  & 93.22  & -0.08   & 2.30   & 15.90   \\
			SFP \cite{DBLP:conf/ijcai/HeKDFY18A51}         & 93.68  & 93.93  & -0.25   & $-$    & 28.20   \\
			Li et al. \cite{DBLP:conf/iclr/0022KDSG17A15}   & 93.53  & 93.30  &  0.20  & 32.40  & 38.60   \\
			Liu et al. \cite{DBLP:conf/iclr/LiuSZHD19A50}  & 93.14  & 93.60  & -0.46   & 32.40  & 38.60   \\
			HRank \cite{DBLP:conf/cvpr/LinJWZZ0020A30}       & 93.50  & 94.23  & -0.73 & 41.20  & 39.40   \\
			SFP \cite{DBLP:conf/ijcai/HeKDFY18A51}         & 93.68  & 93.86  & -0.18   & $-$    & 40.80   \\
			CNN-FCF \cite{DBLP:conf/cvpr/LiWYFZL19A54}     & 93.58  & 93.67  & -0.09   & 43.19  & 43.08   \\
			NISP \cite{DBLP:conf/cvpr/Yu00LMHGLD18A17}        & $-$    & $-$    &  0.18    & 43.25  & 43.78   \\
			GAL \cite{DBLP:conf/cvpr/LinJYZCYHD19A58}         & 93.50  & 92.74  &  0.76 & 44.80  & 48.50   \\
			FPGM \cite{DBLP:conf/cvpr/HeLWHY19A29}         & 93.68  & 93.73  & -0.05  & $-$  & 52.30   \\
			Y.He et al. \cite{DBLP:conf/cvpr/HeDLZZ020A41} & 93.68  & 93.79  & -0.11  & $-$  & 60.30   \\
			\textbf{ACP($\varepsilon$=0.085)}  & \textbf{93.32}  & \textbf{94.33}  & \textbf{-1.01}  & \textbf{62.43} & \textbf{63.95}   \\
			\hline
			\hline		
			ABCPruner \cite{DBLP:conf/ijcai/LinJZZW020A57}   & 93.50  & 93.58  & -0.08  & 67.41  & 65.04   \\
			CNN-FCF \cite{DBLP:conf/cvpr/LiWYFZL19A54}     & 93.58  & 92.96  &  0.62   & 69.51   & 70.81   \\
			HRank \cite{DBLP:conf/cvpr/LinJWZZ0020A30}       & 93.50  & 92.65  &  0.85   & 68.60   & 68.70   \\
			\textbf{ACP($\varepsilon$=0.300)}  & \textbf{93.32}  & \textbf{93.65}  & \textbf{-0.33}   & \textbf{80.92}   & \textbf{78.32}   \\			
			\hline 			
	\end{tabular}}	
	\label{tab5}	
\end{table}

\subsubsection{Comparison on CIFAR-10}

First of all, we compare the methods of pruning ResNet-56 on CIFAR-10. The results are listed in TABLE \ref{tab4}. As we can see, our proposed ACP is superior to the existing methods including the state-of-the-art in terms of parameter and FLOPs compressing rate and the performance after pruning. We achieve even 0.21\% accuracy rise with discarding 58.82\% parameters and 54.42\% FLOPs, while the accuracy of the compact network in ABCPruner \cite{DBLP:conf/ijcai/LinJZZW020A57} reduces by 0.03\% with a similar compressing rate (54.20\% versus 58.82\% and 54.13\% versus 54.42\%). Although the performance of the compressed ResNet-56 after pruning by CNN-FCF \cite{DBLP:conf/cvpr/LiWYFZL19A54} improves by 0.24\% which is 0.03\% better than our ACP, the pruning rate of parameters and FLOPs are 15.73\% and 11.64\% lower than ours respectively (43.09\% versus 58.82\% and 42.78\% versus 54.42\%).

Secondly, we compare the pruning methods for ResNet-110 on CIFAR-10. The results are shown in TABLE \ref{tab5}. When we remove 62.43\% parameters and 63.95\% FLOPs, the accuracy of the compact network is 1.01\% higher than the vanilla network, which is far better than other methods. Among the compared schemes, Liu et al. \cite{DBLP:conf/iclr/LiuSZHD19A50} achieve the best performance with 0.46\% accuracy raise when the parameters and FLOPs are pruned by 32.40\% and 38.60\% respectively. When the compressing rate increases to about 80\%, the compact network obtained via ACP still has better performance than the original network and the compared approaches. The results further verify the effectiveness of the proposed method for compressing and accelerating CNN.

\begin{table}[t]	
	\centering
	\caption{Performance comparison of ResNet-56 on CIFAR-100.}
	\renewcommand{\arraystretch}{1.3}
	\resizebox{85mm}{!}{
		\begin{tabular}{c c c c c c}			
			\hline 
			Method  &Baseline Acc/\% & Pruned Acc/\% & Acc.drop/\%  & FLOPs.drop/\% \\ 
			\hline
			Y.He et al. \cite{DBLP:conf/cvpr/HeDLZZ020A41} & 71.41  & 70.83  &  0.58    & 51.60   \\
			SFP \cite{DBLP:conf/ijcai/HeKDFY18A51}         & 71.40  & 68.70  &  2.61    & 52.60   \\
			FPGM \cite{DBLP:conf/cvpr/HeLWHY19A29}         & 71.41  & 69.66  &  1.75    & 52.60   \\
			\textbf{ACP($\varepsilon$=0.060)}  & \textbf{71.36}  & \textbf{71.15}  & \textbf{0.21}  & \textbf{52.21}   \\			
			\hline 			
	\end{tabular}}	
	\label{tab6}	
\end{table}

\begin{table}[t]	
	\centering
	\caption{Performance comparison of ResNet-34 on ILSVRC-2012.}
	\renewcommand{\arraystretch}{1.3}
	\resizebox{85mm}{!}{
		\begin{tabular}{c c c c c}			
			\hline 
			Method  & Top-1 Acc.drop/\% & Top-5 Acc.drop/\% & Parameters.drop/\% & FLOPs.drop/\% \\ 
			\hline
			Li et al. \cite{DBLP:conf/iclr/0022KDSG17A15}   & 0.75  & $-$   & 7.20    & 7.50     \\
			Liu et al. \cite{DBLP:conf/iclr/LiuSZHD19A50}  & 0.40  & $-$   & 10.80   & 24.20    \\
			Taylor-FO-BN \cite{DBLP:conf/cvpr/MolchanovMTFK19A56} & 0.48  & $-$   & $-$     & 24.20    \\
			\textbf{ACP($\varepsilon$=0.010)}  & \textbf{0.37}  & \textbf{0.12}  & \textbf{29.45}  & \textbf{31.22}   \\
			\hline
			\hline	
			FPGM \cite{DBLP:conf/cvpr/HeLWHY19A29}        & 1.29  & 0.54   & $-$     & 41.10     \\
			SFP \cite{DBLP:conf/ijcai/HeKDFY18A51}         & 2.09  & 1.29   & $-$     & 41.10     \\
			NISP \cite{DBLP:conf/cvpr/Yu00LMHGLD18A17}        & 0.92  & $-$    & 43.68   & 43.76     \\
			EDP \cite{9246734A53}         & 1.13  & 0.51   & 45.50   & 44.90     \\
			ABCPruner \cite{DBLP:conf/ijcai/LinJZZW020A57}   & 2.30  & 1.40   & 53.58   & 41.00     \\
			CNN-FCF \cite{DBLP:conf/cvpr/LiWYFZL19A54}     & 1.97  & 1.22   & 55.80   & 54.87     \\	
			
			\textbf{ACP($\varepsilon$=0.030)}  & \textbf{1.32}  & \textbf{0.87}  & \textbf{58.07}  & \textbf{55.73}   \\			
			\hline 			
	\end{tabular}}	
	\label{tab7}	
\end{table}

\subsubsection{Comparison on CIFAR-100}
In the CIFAR-100 experiments, we tabulate the compared results of pruning ResNet-56 in TABLE \ref{tab6}. The original accuracy is only 71.36\% due to the inadequate training set. In this case, the redundancy of the network is relatively scanty, and it also enhances the difficulty of compressing. Although the pruned accuracy drops to an extent compared to the original network, our ACP obtains better parameters and FLOPs reduction and performance than other methods. SFP \cite{DBLP:conf/ijcai/HeKDFY18A51} and FPGM \cite{DBLP:conf/cvpr/HeLWHY19A29} remove 52.60\% FLOPs with 2.61\% and 1.75\% accuracy loss, respectively. However, at the nearly same FLOPs pruning rate with ACP (52.21\% versus 52.60\%), the accuracy only degrades by 0.21\%.

\subsubsection{Comparison on ILSVRC-2012}

In the end, we compare the performance of the pruning methods of ResNet-34 and ResNet-50 on ILSVRC-2012. The results are depicted in TABLE \ref{tab7} and TABLE \ref{tab8}. It can be found from TABLE \ref{tab7} that the performance of ACP is generally superior among compared methods. Although the Top-1 and Top-5 accuracy loss is slightly higher than that of NISP \cite{DBLP:conf/cvpr/Yu00LMHGLD18A17}, FPGM \cite{DBLP:conf/cvpr/HeLWHY19A29}, and EDP \cite{9246734A53}, the pruning rate of parameters and FLOPs of ACP is over 10\% higher than the three approaches. When we discard 29.45\% parameter and 31.22\% FLOPs of ResNet-34, the performance of the compact network is better than Li et al. \cite{ISI:000450913101044A13}, Liu et al. \cite{DBLP:conf/iclr/LiuSZHD19A50}, and Taylor-FO-BN \cite{DBLP:conf/cvpr/MolchanovMTFK19A56}.

As we can see from TABLE \ref{tab8}, the performance of ACP significantly exceeds the other schemes. Although the Top-5 accuracy loss of the pruned network using ACP is 0.02\% (0.21\% versus 0.19\%) higher than that of CNN-FCF \cite{DBLP:conf/cvpr/LiWYFZL19A54}, the Top-1 accuracy loss reduces by 0.06\% (0.41\% versus 0.47\%). Moreover, the compressing rate of the parameters increases by 12.28\% compared with CNN-FCF \cite{DBLP:conf/cvpr/LiWYFZL19A54} in the case of nearly the same pruning rate of FLOPs. The Top-5 accuracy loss of ThiNet \cite{DBLP:journals/pami/LuoZZXWL19A59} is 0.12\% less than ACP, but its Top-1 accuracy loss is even 0.86\% higher with 20.97\% (33.72\% versus 54.69\%) and 10.03\% (36.79\% versus 46.82\%) lower parameters and FLOPs compressing rate respectively. Additionally, as the pruning rate increases, the dominance of ACP becomes more obvious. All the results manifest that the proposed ACP still has a good performance on large-scale image classification tasks.

\begin{table}[t]	
	\centering
	\caption{Performance comparison of ResNet-50 on ILSVRC-2012.}
	\renewcommand{\arraystretch}{1.3}
	%\resizebox{\textwidth}{!}
	\resizebox{85mm}{!}{
		\begin{tabular}{c c c c c}			
			\hline 
			Method  & Top-1 Acc.drop/\% & Top-5 Acc.drop/\% & Parameters.drop/\% & FLOPs.drop/\% \\ 
			\hline
			ThiNet       & 1.27  & 0.09   & 33.72   & 36.79     \\
			SFP \cite{DBLP:conf/ijcai/HeKDFY18A51}         & 1.54  & 0.81   & $-$     & 41.80     \\
			HRank \cite{DBLP:conf/cvpr/LinJWZZ0020A30}       & 1.17  & 0.54   & 36.67   & 43.77     \\
			NISP \cite{DBLP:conf/cvpr/Yu00LMHGLD18A17}        & 0.89  & $-$    & 43.82   & 44.01     \\
			Taylor-FO-BN \cite{DBLP:conf/cvpr/MolchanovMTFK19A56} & 1.68  & $-$    & $-$     & 45.00     \\
			CNN-FCF \cite{DBLP:conf/cvpr/LiWYFZL19A54}      & 0.47  & 0.19   & 42.41   & 46.05     \\
			EDP \cite{9246734A53}         & 0.56  & 0.34   & 43.90   & 52.60     \\
			\textbf{ACP($\varepsilon$=0.020)}  & \textbf{0.41}  & \textbf{0.21}  & \textbf{54.69}  & \textbf{46.82}   \\
			\hline
			\hline
			He et al. \cite{ISI:000425498401048A14}	 & $-$   & 1.40   & $-$     & 50.00     \\
			FPGM \cite{DBLP:conf/cvpr/HeLWHY19A29}        & 1.32  & 0.55   & $-$     & 53.50     \\
			CCP \cite{DBLP:conf/icml/PengWCH19A55}         & 0.94  & 0.45   & $-$     & 54.10     \\
			GAL \cite{DBLP:conf/cvpr/LinJYZCYHD19A58}         & 4.35  & 2.05   & $-$     & 55.00     \\
			ABCPruner \cite{DBLP:conf/ijcai/LinJZZW020A57}   & 2.49  & 1.45   & 56.01   & 56.61     \\
			CNN-FCF \cite{DBLP:conf/cvpr/LiWYFZL19A54}     & 1.60  & 0.69   & 52.52   & 57.10     \\
			Y.He et al. \cite{DBLP:conf/cvpr/HeDLZZ020A41}  & 1.97  & 0.95   & $-$     & 60.80     \\
			HRank \cite{DBLP:conf/cvpr/LinJWZZ0020A30}       & 4.17  & 1.86   & 46.00   & 62.10     \\				
			\textbf{ACP($\varepsilon$=0.040)}  & \textbf{0.89}  & \textbf{0.53}  & \textbf{64.36}  & \textbf{63.34}   \\			
			\hline 			
		\end{tabular}
	}	
	\label{tab8}	
\end{table}

\begin{figure}[t]
	
	\centering
	
	\includegraphics[width=8.5cm]{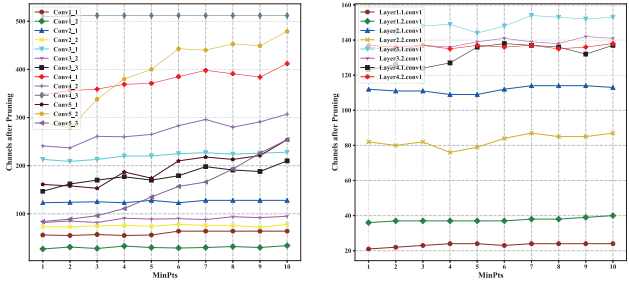} 
	
	\caption{The influence of \textit{MinPts} on the channels. (\textbf{Left}) Pruning VGG-16 on CIFAR-10, where $\varepsilon$=0.010; (\textbf{Right}) Pruning Resnet-18 on ILSVRC-2012, where $\varepsilon$=0.010.}
	\label{img4}
	
\end{figure}

\subsection{Ablation Analysis}

Then, we conduct ablation analysis on the proposed ACP method. This section is composed by six parts: the influence of $MinPts$, the influence of $\varepsilon$, the influence of different similarity criteria, the influence of PSO, the effectiveness of ACP and generalization ability on detection tasks.

\subsubsection{The Influence of $MinPts$}

$MinPts$ is a hyper-parameter for clustering in our method. We analyze the influence of $MinPts$ on the number of pruned channels in the clustering pruning stage. The results are displayed in Fig.\ref{img4}. Given the pre-trained model, we set $\varepsilon$=0.010 and change $MinPts$ from 1 to 10 to record the number of remaining channels in each layer after pruning. The left image is the results of pruning VGG-16 on CIFAR-10. It can be seen that only the last two layers Conv5\_2 and Conv5\_3 significantly fluctuate with $MinPts$, and the other layers are relatively stable or only have slight variances. The number of channels in more than half layers keeps almost unchanged after compressing with the change of $MinPts$. The right image is the results of pruning Resnet-18 on ILSVRC-2012. It manifests that the number of pruned channels in all layers varies with $MinPts$ to a small extent. Therefore, controlling the pruning rate by changing $MinPts$ will hamper the flexibility of the network compression.

\begin{figure}[t]
	
	\centering
	
	\includegraphics[width=8.5cm]{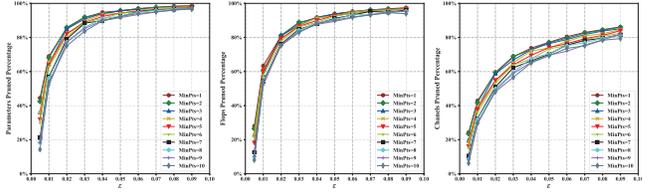} 
	
	\caption{The influence of ($\varepsilon$, \textit{MinPts}) for the pruned percentage of parameters, FLOPs and channels of VGG-16 on CIFAR-10.}
	\label{img5}
	
\end{figure}

In order to select a better $MinPts$, we perform experiments on the effect of different hyper-parameter combinations ($\varepsilon$, $MinPts$) on the number of parameters, FLOPs, and channels when compressing VGG-16 on CIFAR-10 in the clustering pruning phase. The experimental results are shown in Fig.\ref{img5}. When $MinPts$=5 (the red curve), the pruning rates of the original network change more smoothly with variable $\varepsilon$ than the other values of $MinPts$, meanwhile the coverage of all pruning rates is also relatively moderate. Comprehensively considering the adjustment of hyper-parameters and the compressing rate of the network, we make $MinPts$=5 unchanged and only change $\varepsilon$ to perform our ACP. The experimental results in the previous section entirely support the effectiveness of this hyper-parameter setting method.

\begin{figure}[t]
	
	\centering
	
	\includegraphics[width=8.5cm]{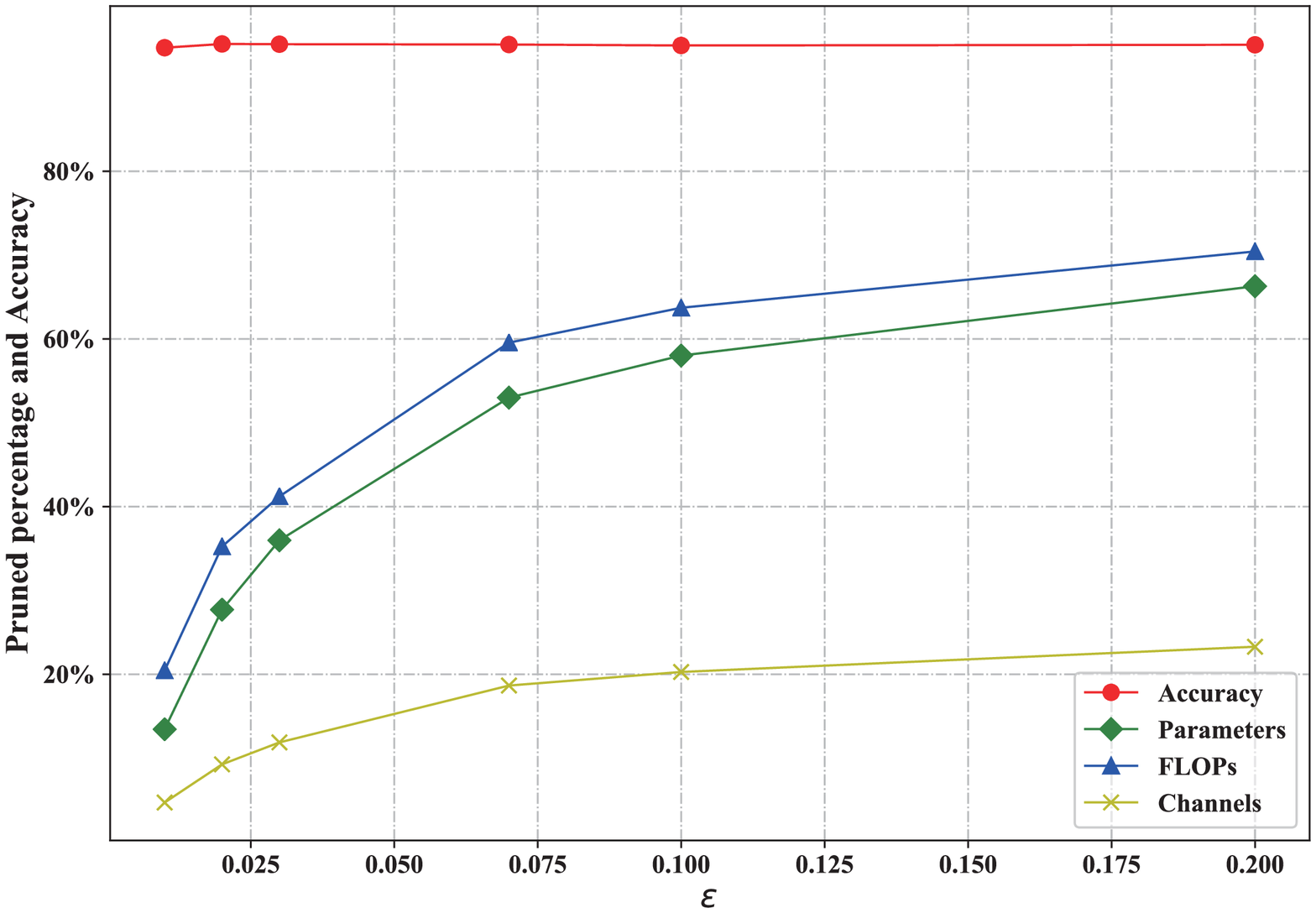} 
	
	\caption{The influence of $\varepsilon$ for the pruned percentage and accuracy of GoogLeNet on CIFAR-10.}
	\label{img6}
	
\end{figure}

\begin{figure*}[htbp]
	
	\centering
	
	\includegraphics[width=17cm]{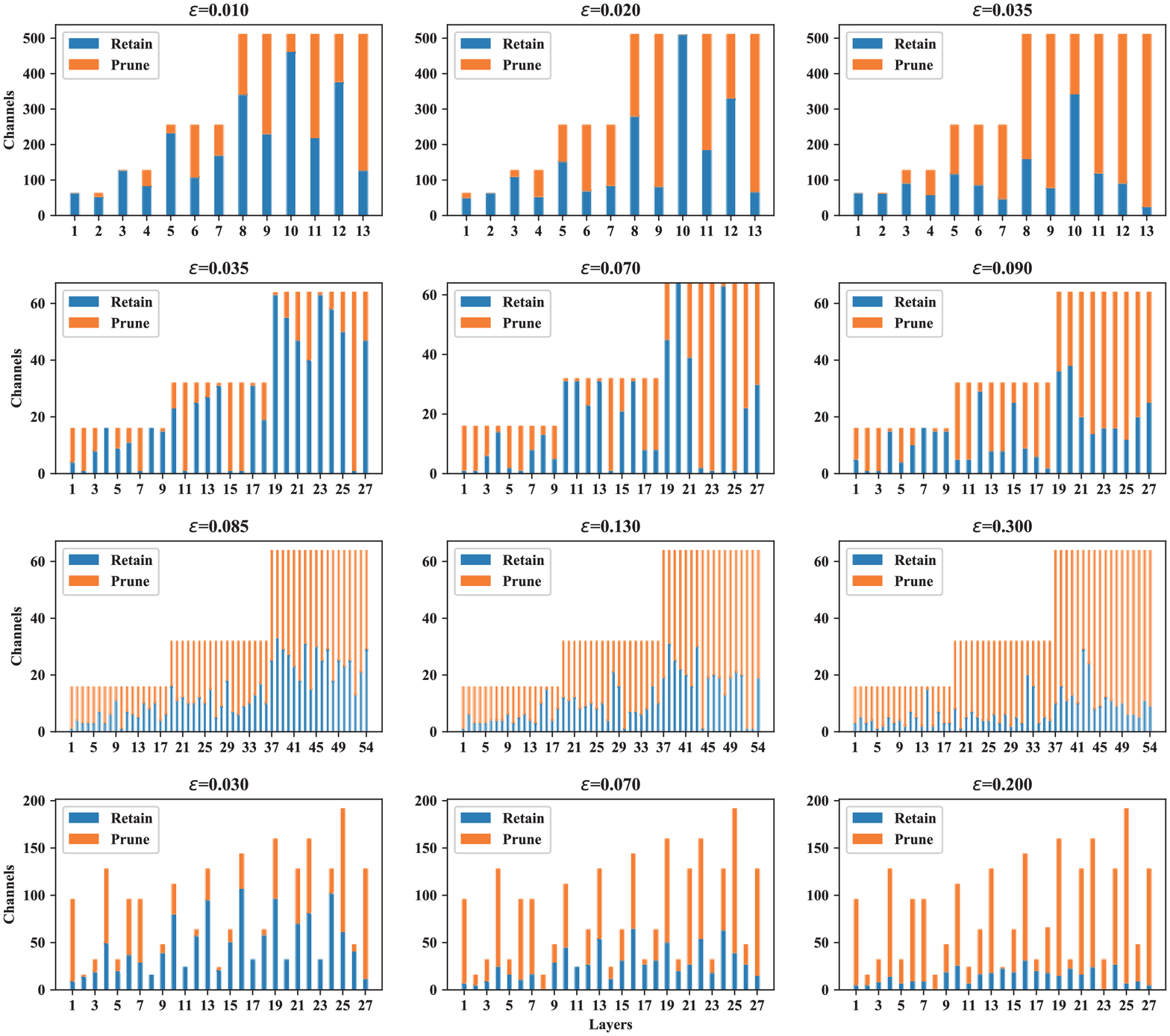} 
	
	\caption{The influence of $\varepsilon$ for the pruned percentage and accuracy of GoogLeNet on CIFAR-10.}
	\label{img7}
	
\end{figure*}

\begin{table*}[t]	
	\centering
	\caption{Performance comparison of pruned networks with and without PSO. $``$Acc.w.o.$"$ means we compress networks only with the clustering pruning; $``$Acc.w.$"$ means we optimize compact networks after clustering pruning via PSO.}
	\renewcommand{\arraystretch}{1.3}
	{
		\begin{tabular}{c c c c c c c c c c c c}			
			\hline 
			Dataset  & \multicolumn{9}{c}{CIFAR-10} & \multicolumn{2}{c}{CIFAR-100} \\ 
			\hline
			Model    & \multicolumn{3}{c}{VGG-16}  & \multicolumn{3}{c}{ResNet-56}  &  \multicolumn{3}{c}{GoogLeNet}  &  \multicolumn{2}{c}{ResNet-56} \\
			
			\cmidrule(r){1-1}  \cmidrule(r){2-4}  \cmidrule(r){5-7} \cmidrule(r){8-10} \cmidrule(r){11-12}
			
			$\varepsilon$  &  0.010  &  0.020  &  0.035  &  0.035  &  0.075  &  0.080  &  0.030  &  0.070  &  0.200  &  0.030  &  0.060  \\
			Baseline Acc/\%    & \multicolumn{3}{c}{93.60}  & \multicolumn{3}{c}{93.18}  &  \multicolumn{3}{c}{94.72}  &  \multicolumn{2}{c}{71.36} \\
			Acc.w.o./\%  &  93.56  &  93.21  &  91.82  &  \textbf{93.81}  &  93.15  &  \textbf{93.53}  &  95.00  &  95.10  &  94.79  &  70.32  &  70.50  \\
			Acc.w./\%   &  \textbf{94.03}  &  \textbf{93.66}  &  \textbf{93.45}  &  93.78  &  \textbf{93.39}  &  92.91  &  \textbf{95.16}  &  \textbf{95.12}  &  \textbf{95.09}  &  \textbf{71.77}  &  \textbf{71.15} \\		    
			\hline 			
	\end{tabular}}	
	\label{tab9}	
\end{table*}

\subsubsection{The Influence of $\varepsilon$}

The neighborhood radius $\varepsilon$ is the only one hyper-parameter in our method. Since $\varepsilon$ can restrict the size of the clusters formed in the clustering pruning, we adjust it to control the pruning rate of the CNN. The larger the $\varepsilon$, the fewer the channel clusters in each layer, and the greater the degree of network compression. Under this circumstance, the number of channels, parameters, and FLOPs will decrease, and the accuracy of the compact network will proceed to reduce. Here, we compress GoogLeNet on the CIFAR-10 to verify the influence of $\varepsilon$ for network pruning. The results are shown in Fig.\ref{img6}. From the figure, it is clear that when the compressing rate is tiny, the pruning rate of three indicators changes more obviously, and as the compression amplitude increases, the changes gradually become gentle. This is because when $\varepsilon$ is small, the clusters are more sensitive, and a slight raise of $\varepsilon$ has a great impact on the clustering results and the pruning rate. With the increase of $\varepsilon$, the clustering tends to be stable, and then increasing $\varepsilon$ will have a small effect on the network pruning rate.

Fig.\ref{img7} presents the influence of $\varepsilon$ on the number of channels before and after pruning on CIFAR-10. From top to bottom are VGG-16, ResNet-56, ResNet-110, and GoogLeNet. As $\varepsilon$ increases, the network compressing rate gradually raises. As can be seen, not all the number of channels in each layer gradually declines with the increase of the network pruning rate. For example, when $\varepsilon$=0.020, the pruning rate of the tenth layer of VGG-16 is lower than that with $\varepsilon$=0.010. It reveals that when the overall compressing rate of the network raises, increasing the number of channels in several layers can even improve the performance of the compact network. These layers are more important for feature extraction in image classification tasks when the network is compressed remarkably. Meanwhile, it further demonstrates that the ACP can find an optimal compact structure with superior performance according to different pruning rates.

\subsubsection{The Influence of Different Similarity Criteria}

\begin{figure}[t]
	
	\centering
	
	\includegraphics[width=8.5cm]{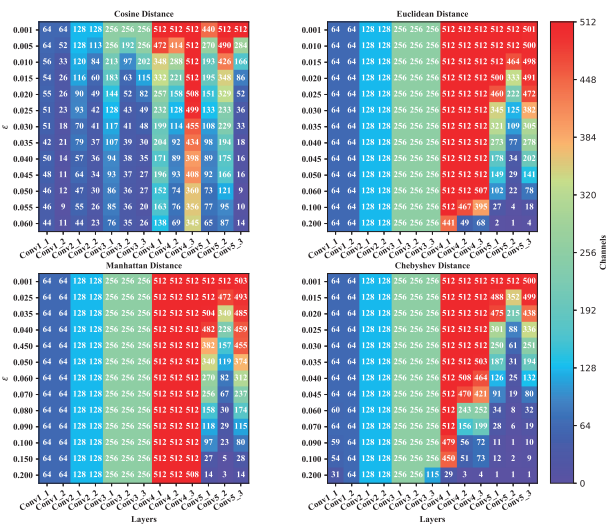} 
	
	\caption{The influence of different similarity criteria for clustering pruning of VGG-16 on CIFAR-10.}
	\label{img8}
	
\end{figure}

We use cosine distance to measure the similarity of the feature maps and guide clustering. In this section, we continue to estimate other similarity measurements for feature maps. Specifically, the Cosine Distance, Euclidean Distance, Manhattan Distance, and Chebyshev Distance are selected to cluster and prune VGG-16 on CIFAR-10. We analyze the impact of four similarity on clustering pruning and the experimental results are depicted in Fig.\ref{img8}. When we use Cosine Distance, the compression rate of the network gradually increases as $\varepsilon$ raises, meanwhile the number of channels in all layers constantly decreases. While Euclidean Distance is applied, the number of channels of the first seven layers remains unchanged with pruning, only that of the last three layers reduce. As the last three layers are almost discarded to one channel, the previous seven layers are still no cutting. For Manhattan Distance, only the number of channels of the last three layers keep decreasing, while the first nine layers are unchanged. Finally, as for Chebyshev Distance, the changes in the last six layers are more obvious, but the first seven layers are still almost unpruned. We conjecture this is because the dimension of the feature maps closed to the input is higher than that closed to the output. When the Cosine Distance is adopted to measure the similarity of feature maps, it will not be affected by their dimension. Even for high-dimension features, the Cosine Distance is still accurate and effective. The other three kinds of similarity are less sensitive to the dimension of features. This experiment entirely proves the effectiveness of Cosine Distance in the similarity measure of feature maps and the flexibility for channel pruning.

\subsubsection{The Influence of PSO}

To verify the effect of PSO, we further compare the performance of the compact network with and without optimization. As can be seen from TABLE \ref{tab9}, when ResNet-56 is compressed on CIFAR-10 with $\varepsilon$=0.035, the accuracy without optimizing is 0.03\% higher than that with optimizing, and when $\varepsilon$ is 0.080, the accuracy is 0.62\% higher without optimization. In the other nine groups of experiments, the performance with swarm intelligence searching is better than that by directly clustering pruning. On CIFAR-10, when $\varepsilon$ is 0.035, the accuracy of the compressed VGG-16 with PSO is even 1.63\% higher than that without optimization. This experiment adequately verifies the rationality of the proposed automatic channel pruning scheme based on clustering and swarm intelligence optimization. Although, the performance of the subnetwork obtained by clustering pruning may be poor, it shrinks the initializing and searching space of the structure candidate and relieves the optimizing pressure of swarm intelligence optimization. Finally, after several iterations, the performance of the compact network will be significantly improved, which is close to the baseline or even better.

\subsubsection{The Effectiveness of ACP}

In this section, we compare the proposed ACP to the following six methods Slimming \cite{ISI:000425498402086A26}, WLP \cite{ISI:000450913101044A13}, He et al. \cite{ISI:000425498401048A14}, SFP \cite{DBLP:conf/ijcai/HeKDFY18A51}, HRank \cite{DBLP:conf/cvpr/LinJWZZ0020A30}, and ABCPruner \cite{DBLP:conf/ijcai/LinJZZW020A57}. Among them, Slimming performs sparse induction regularization on the scaling factor of the BN layer and then prunes the channels with smaller scaling factors according to the compressing ratio. WLP trims unimportant connections with smaller weights according to a preset trimming rate. He et al. utilize a channel selection method based on LASSO regression to delete redundant channels. SFP applies the L2-norm to evaluate the importance of filters and discards the useless filters and related channels. HRank adopts the rank of the feature maps to evaluate the significance of the channels and then compresses the network. The above five methods are all structured pruning methods. ABCPruner formulates the network compression as an optimization problem to find the optimal pruned structure via artificial bee colony algorithm.

\begin{figure}[t]
	
	\centering
	
	\includegraphics[width=8.5cm]{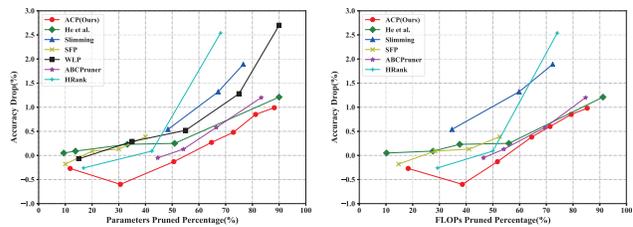} 
	
	\caption{Comparison of different pruning methods for ResNet-56 on CIFAR-10.}
	\label{img9}
	
\end{figure}

In order to achieve a fair comparison, we prune ResNet-56 on CIFAR-10 by the above methods with the same environment and parameter settings on a 8G GTX 1070 GPU. The accuracy loss of each pruning scheme with different parameters and FLOPs pruned percentage is depicted in Fig.\ref{img9}. In general, as the network compression rate raises, the precision of the compact network obtained by all seven approaches trends to decline. When the pruning rate of parameters and FLOPs reaches 30\% and 40\% respectively with our ACP, the pruned performance is best, which even 0.5\% higher than the vanilla network. Both ABCPruner and our ACP draw on the idea of neural structure search, and finally, the performance is better than the other five structured pruning methods. It unveils that neural structure search can optimize and promote the network compressing. With compressing, the accuracy drop of the pruned network via ACP is always less than the other methods. The results validate that the automatic channel pruning is cutting-edge in compression and accelerating CNNs.

\subsubsection{Generalization Ability on Detection Tasks}

As described above, ACP is effective in image classification. To further analyze the generalization of our method, we conduct experiments on object detection. Specially, we use pruned VGG-16 on CIFAR-10 as a backbone network to deploy SSD on PASCAL VOC. The training images of PASCAL VOC 2007 and 2012 are combined for training, and then the 4963 testing images in PASCAL VOC 2007 are used to estimate the performance. Here, we tabulate the pruning rate of parameters and FLOPs of SSD and the loss of Mean Average Precision (mAP) in TABLE \ref{tab10}. The result indicates that SSD has significant redundancy on PASCAL VOC. When discarding 35.41\% parameters and 35.85\% FLOPs respectively, the mAP is even increased by 0.79\%. Furthermore, pruning 61.13\% parameters and 66.44\% FLOPs slightly changes performance with 0.40\% mAP value drop but significantly decreases storage and computational requirements. This experiment proves the our ACP also has good generalization on object detection.

\section{Conclusion}
In this work, we propose a novel automatic channel pruning method for compressing and accelerating CNNs. We use the cosine distance between feature maps to perform clustering and pruning for channels. Additionally, the PSO algorithm is introduced to iteratively search for the optimal compact network based on the candidate population initialized by the clustering pruned substructure. The compressed model is then retrained to mitigate the accuracy loss from pruning. We also analyze the impact of various similarities on the clustering and demonstrate the cosine distance can be applied to measure the similarity for feature maps of different dimensions. Extensive experiments have manifested that ACP is comparable with state-of-the-art pruning methods in performance and pruning rate of parameters and FLOPs. On CIFAR-10, when discarding more than 50\% parameters and FLOPs of VGGNet, ResNet, and GoogLeNet, their performance even improves. Our proposed method achieves 60\% compression with only 0.89\% and 0.53\% drop in Top-1 and Top-5 accuracy respectively for ResNet-50 on ILSVRC-2012. Moreover, ablation analysis has shown that ACP is generally applicable to other real world tasks and networks.

In the future, we will integrate the channel pruning method with other compression schemes such as quantization. Furthermore, we will consider applying existing approaches to accelerate other real-world vision tasks and even natural language processing.

\begin{table}[t]	
	\centering
	\caption{Pruning SSD on PASCAL VOC for object detection.}
	\renewcommand{\arraystretch}{1.3}
	\resizebox{85mm}{!}{
		\begin{tabular}{c c c c c}			
			\hline 
			Method  & mAP/\% & mAP.drop/\% & Parameters.drop/\% & FLOPs.drop/\% \\ 
			\hline
			Baseline                      & 74.13  & $-$     & $-$     & $-$      \\
			ACP($\varepsilon$=0.040)      & 74.92  & -0.79   & 35.41   & 35.85     \\
			ACP($\varepsilon$=0.040)      & 74.80  & -0.67   & 50.55   & 49.51     \\
			ACP($\varepsilon$=0.040)      & 73.73  & 0.40    & 61.13   & 66.44     \\	
			\hline					
		\end{tabular}
	}	
	\label{tab10}	
\end{table}

% or


\section*{Acknowledgment}

This work was supported in part by the National Key Research and Development Program under Grant 2018YFC0604404, in part by the National Natural Science Foundation of China under Grant 61806067, and in part by the Anhui Provincial Key R\&D Program (202004a05020040).

%\ifCLASSOPTIONcaptionsoff
  %\newpage
%\fi



%\bibliographystyle{IEEEtran}

\bibliography{mybib}

\end{document}